\documentclass[journal]{IEEEtran}
\usepackage{amsmath,amsfonts}
\usepackage{algorithm}
\usepackage{algorithmicx}
\usepackage{algpseudocode}
\usepackage{array}
\usepackage[caption=false,font=normalsize,labelfont=sf,textfont=sf]{subfig}
\usepackage{textcomp}
\usepackage{stfloats}
\usepackage{url}
\usepackage{verbatim}
\usepackage{graphicx}
\usepackage{amsmath}
\usepackage{caption}
\usepackage[sorting=none,maxnames=4,maxbibnames=4,style=ieee]{biblatex}
\usepackage{tabularx}
\usepackage{amsmath, amssymb}

\begin{document}

\title{Bayesian Neural Networks: A Min-Max Game Framework}

\author{Junping Hong, Ercan Engin Kuruoglu,~\IEEEmembership{Senior Member,~IEEE,}

\thanks{The authors are with Tsinghua-Berkeley Shenzhen Institute, Shenzhen International Graduate School, Tsinghua University, Shenzhen, China, 518000 (emails: jacob.hong17@gmail.com; kuruoglu@sz.tsinghua.edu.cn).}}

\markboth{IEEE TRANSACTIONS ON SIGNAL AND INFORMATION PROCESSING OVER NETWORKS,~Vol.~XX, No.~X, August~20XX}%
{Shell \MakeLowercase{\textit{et al.}}: A Sample Article Using IEEEtran.cls for IEEE Journals}

\IEEEoverridecommandlockouts
\IEEEpubid{\begin{minipage}[t]{\textwidth}\ \\[8pt]
        \centering\normalsize{0000--0000/00\$00.00 \copyright~2021 IEEE}
\end{minipage}} 


\maketitle

\begin{abstract}
In deep learning, Bayesian neural networks (BNN) provide the role of robustness analysis, and the minimax method is used to be a conservative choice in the traditional Bayesian field. In this paper, we study a conservative BNN with the minimax method and formulate a two-player game between a deterministic neural network $f$ and a sampling stochastic neural network $f + r*\xi$. From this perspective, we understand the closed-loop neural networks with the minimax loss and reveal their connection to the BNN. We test the models on simple data sets, study their robustness under noise perturbation, and report some issues for searching $r$. 
\end{abstract}


\section{Introduction}
\IEEEPARstart{N}{owdays}, deep learning, as a data-driven method, has become more and more popular and has been applied to multiple areas, such as weather forecasting \cite{bi2022pangu}, large language models \cite{wei2022emergent}, image classification \cite{li2019deep,kuruoglu2010using}. Most neural networks are trained with supervised learning with an end-to-end framework. Representation learning seeks a good representation of the trained data, such as learning representation by mutual information (MI) \cite{hjelm2018learning}, and maximal coding reduction (MCR), which can be viewed as a nonlinear principal component analysis \cite{yu2020learning}.  

Although deep learning seems to be rather successful, robustness issues still haunt the society of deep learning \cite{liu2018analyzing, fawzi2017robustness}. \cite{mackay1992practical} proposed the issues about determinist neural networks and introduced the first Bayesian framework in neural networks, and \cite{neal2012bayesian} made a further study. Dropout as a well-known technique is proposed to prevent overfitting \cite{srivastava2014dropout} and can be seen as an approximation of Bayesian \cite{gal2016dropout}. BNN aims to learn a distribution of neural networks through posterior estimation and use random variables to describe the weights of neural networks and update the mean and the variance simultaneously \cite{jospin2022hands}. Previous works have shown that BNN can quantify the uncertainty of neural networks \cite{blundell2015weight}, is robust to the choice of prior \cite{izmailov2021bayesian}, and is more robust to gradient attack than deterministic neural networks \cite{carbone2020robustness}. Prior Networks \cite{malinin2018predictive}, the authors argue that the randomness of deep learning includes model uncertainty, data uncertainty, and distributional uncertainty, and utilizes the Prior Networks to do the out-of-distribution detection. \cite{wang2021pac} applied the MCMC methods to the information bottleneck study.  In addition, the minimax method is often thought of as a robustness help for Bayesian methods \cite{berger2013statistical}, and it will improve the robustness at the cost of accuracy because it considers the best case of the worst case.  

The minimax method has been used for a long time. Previous studies using the minimax game to study the robustness of neural networks are the fault-tolerant neural networks \cite{neti1992maximally,deodhare1998synthesis,duddu2019adversarial}, obtain the two-player game between normal neural network and the fault neural networks. The closed-loop transcription neural networks \cite{dai2022ctrl,dai2023closed}, design a new two-player game between the decoder and composition of encoder and decoder with minimax coding rate reduction. Besides,  people also apply the minimax game with Bayesian consideration in reinforcement learning \cite{buening2023minimax}.

Inspired by the minimax work in the representation level \cite{dai2023closed} and the classical BNN, we try to apply the minimax game to the classical BNN. The reason is to understand the closed-loop neural networks and reveal their connection with BNN, which can also help us to know more about the classical BNN. To the best of our knowledge, this is the first time the minimax method has been applied to BNN. Similar works are the fault-tolerant neural networks \cite{duddu2019adversarial} with two differences. One is that we use perturbation rather than fault nodes or edges. The other is that we care about the representation level. Compared with the closed-loop transcription networks \cite{dai2023closed}, they introduce another deterministic neural network $g$ to obtain the closed-loop transcription networks. However, we use a stochastic neural network instead. 

The contribution of this paper includes 3 perspectives. First, our framework can clearly explain the closed-loop neural networks with the minimax loss, and uncover their connection to the BNN. Second, this formulation is naturally more conservative compared with the classical BNN and can provide a reference for the variance setting for the classical BNN. Third, this formulation provides an alternative way for the out-of-distribution detection though usually more complicated than \cite{liang2017enhancing}. In addition, we report some tiny issues for implementing the model. 

The paper is organized as follows: 
Section~\ref{section: model} introduces the formulation of the minimax BNN. 
Section~\ref{section: experiments} presents the experiments and results. 
Section~\ref{section: conclusion} concludes.

\section{MinMax Bayesian Neural Networks}
\label{section: model}

First is the minimax loss in the representation learning, note the loss is the same as \cite{yu2020learning,dai2023closed}, the main difference is that our minimax game is between $f$ and $g=f+r*\xi$. If we view the classical BNN as a hypersphere, then $f$ is the center, and $g$ is the equator.

\begin{equation}
\begin{aligned}
\min \limits_{g} \max \limits_{f} \tau(f,g) \doteq \\ \Delta R(f(X))+\Delta R(g(X)) + \sum_{i=1}^k \Delta R(f(X),g(X)) \\
= \Delta R(Z)+\Delta R(\widehat{Z})+  \sum_{i=1}^k \Delta (R(Z,\widehat{Z})),\label{eq_minimax}
\end{aligned}
\end{equation}

where $X$ denotes the data, $f$ denotes a deterministic neural network, and $g=f+r*\xi$ denotes the sampling stochastic neural network. $\Delta R(f(X))$, $\Delta R(g(X))$, and $\sum_{i=1}^k \Delta R(f(X),g(X))$ denote MCR loss for different case, please see \cite{dai2023closed}. $Z=f(X)$ denotes the final representation learning output for $f$, and $\widehat{Z}=g(X)$ denotes the same for $g$. Note that the third term of this loss controls the gap between $f$ and $g$, and the current setting allows the gap to be pretty large. Hence we will use the $log$ case to denote a smaller gap for $log(1 + \sum_{i=1}^k \Delta (R(Z,\widehat{Z})))$. 

Because MCR loss is not easy to understand, here we provide an equivalent formulation to follow under supervised learning. We do not use this formulation because there exists similar work \cite{duddu2019adversarial}.

\begin{equation}
\begin{aligned}
\min \limits_{f,g} \tau(f,g) \doteq loss(f(X))+ loss(g(X)) \\ \quad \text{s.t.} \quad
\begin{aligned}
|pre(f(X)) - pre(g(X))| = c,
\end{aligned}
\end{aligned}
.\label{eq_minimax_general}
\end{equation}
where $X$, $f$, and $g$ denote the same with previous case. $loss(f(X))$ and $loss(g(X))$ are the loss for $f$ and $g$, such as cross-entropy for classification. $pre(f(X))$ and $pre(g(X))$ are the final prediction for $f$ or $g$. $c$ is a constant that denotes the gap between $f$ and $g$, for example, $c$ could be $100$ for $1000$ predictions, which means we allow the maximal difference between $f$ and $g$ is $10$ percent. Note we can use the Lagrange method to transform this into a min-max or max-min formulation \cite{duddu2019adversarial}.

Compared with the classical BNN, this framework will train two neural networks, one is the mean or the center point, another is the equator, while classical BNN only sample one point. Because MinMax BNN considers both the best case and the worst case, it is more robust than the classical BNN with the cost of performance due to the perturbation in $g$. Another difference is that the variance of weights is given in advance and updated across the training process, while the variance level of MinMax BNN is given by the setting gap and updated by the sampling. Then for the closed-loop neural networks with the same minimax loss in the representation \cite{dai2023closed}, the two deterministic neural networks are $f$ and $f \circ g\circ f$. The main issue is that $f \circ g\circ f$ can not quickly change like $g=f+r*\xi$ since we only need to find a suitable level of $r$. In their formulation, they need to add multiple activation functions to support the image generation, hence they need to use the batch normalization (BN) layer to accelerate \cite{ioffe2015batch}. However, the BN layer seems not suitable for our scenario and will be shown later.   

At the end of this part, we provide a figure to show how radius change affects the minimal process of the loss value with at least a batch of data in Figure 1, where $f$ and the sampling noise $\xi$ are given, and we only need to find the suitable $r$ to obtain $g=f+r*\xi$ to get the minimum point. Normally we should expect the one variable function of $r$ to be concave like Figure 1a and we can search it through the golden search method. However, in many cases, we may not find the correct $r$. One is when the dimension is not enough, we can see the whole function is no longer concave for some sampling $\xi$ like Figure 1b. A similar case will come up when we add bias or BN in the neural networks, a similar case will occur on the tail parts of the function, and their tail is extremely flat compared with Figure 1b. Normally, setting a smaller loss gap and suitable searching zone can reduce this phenomenon, and we can make multiple samples. One last thing is pure linear case can always find the $r$ no matter how large the zone is set, at least with a  \textit{much higher probability}.

\begin{figure}[hpbt]
     \centering
         \centering{
         \includegraphics*[width=3cm]{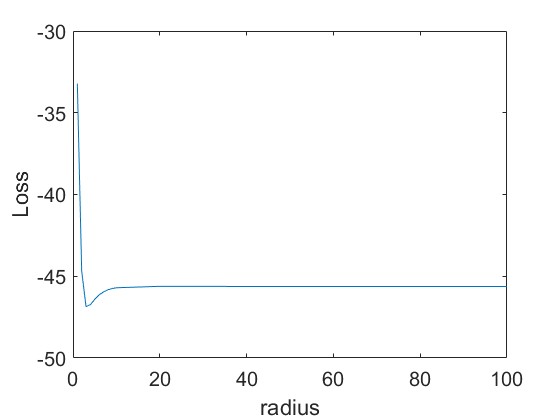}
         \caption*{(a) Normal case, concave function}
         \includegraphics*[width=3cm]{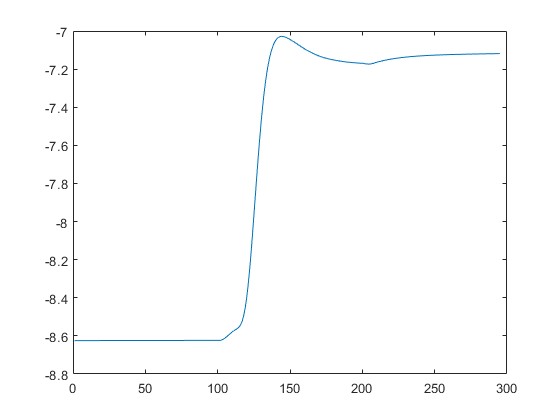}
         \caption*{(b) Strange case without enough dimensions, $log$}
         \includegraphics*[width=3cm]{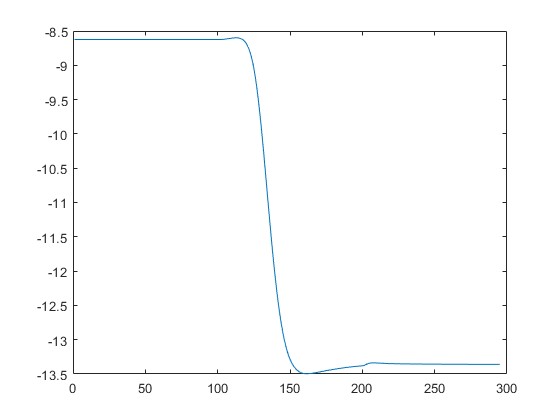}
         \caption*{(c) Strange case without enough dimensions}}
\caption{Representation loss with different radius values or perturbation levels generated by grid search: The interval of 100 points in (b) and (c) is $0.01$, the interval next 100 points is $0.1$, and the interval of the last 100 points is $1$.}
\label{fig_function1}
\end{figure}


\section{Experiments and Analysis}
\label{section: experiments}
The data sets include MNIST \cite{lecun1998mnist}, Fashion MNIST (FMNIST) \cite{xiao2017fashion}, CIFAR-10 \cite{krizhevsky2009learning}. For MNIST and FMNIST, we use the same convolutional neural network (CNN) \cite{lecun1995convolutional} as \cite{dai2023closed}. The optimization algorithm is Adam(0.5,0.999), and the learning rate is 0.001. $f$ only has one kind of activation function with Leakyrelu (Relu and Softplus act the same), and we separate the neural network into three cases. One is without bias or BN, one is with bias, and the last one is with BN. Note that we do not consider adding perturbation on the bias term, one reason is for fair comparison to the three cases, and another is that this is the simplest case, and does not change for the BN layer in perturbation.  $f$ is initialized with $N~(0,0.02)$, and $\xi~N(0,0.01)$ and this might change if we require to update the shape of $\xi$ with Bayes by Backpropagation \cite{blundell2015weight}.    Normally the zone for gold search is from $0$ to $100$, and we will use grid search to validate our correctness. After training, we map the data to the subspace and use the knn methods \cite{guo2003knn} to predict the labels implemented through the scikit-learn package \cite{kramer2016scikit}. In addition, the suitable radius $r$ is from $0.2$ to $0.5$ for the $log$ case on MNIST data, and the $r$ will decrease across the training process, and $r$ is usually about $3$ to $6$ without changing to the third term. We build the model through Pythorch \cite{paszke2019pytorch}, and the codes are public at https://github.com/Jacob-Hong17/MinMax-BNN.

\subsection{Main results}
In Table 1, we can see that the results of MinMax BNN are slightly worse than the closed-loop work \cite{dai2022ctrl} because our formulation is also minimax, and the component neural network will bring more noise in the training process. In Table 2, we can see the best sampling result is for $r=0.2$, hence the classical BNN can use it as a reference.

\begin{table}[!t]
\caption{MinMax BNN results}
\centering
\begin{tabular}{|c||c||c||c|}
\hline
Models & $f$ (MNIST) & closed-loop (MNIST) &  $f$ (FMNIST)\\ 
\hline
test 1 &96.28\% & 97.69\% & 85.82\% \\
\hline
test 2 & 96.43\% & 97.69\% & 85.79\%\\
\hline
test 3 ($log$)&  96.43\% & 97.69\% & 86.21\%\\
\hline
test 4 ($log$)&  96.70\% & 97.69\% & 86.09\%\\
\hline
\end{tabular}
\end{table}

\begin{table}[!t]
\caption{Sample results with different r on MNIST}
\centering
\begin{tabular}{|c||c||c||c||c|}
\hline
r (test 4) & Max & Min & mean & var \\
\hline
0.1 & 97.01\% & 96.93\% & 96.97\% & 5.0e-8\\
\hline
0.2 & 97.07\% & 96.9\% & 96.96\% &1.5e-7\\
\hline
0.5 & 97.03\% & 96.84\% & 96.93\% &2.6e-7\\
\hline
1 & 96.95\% & 96.65\% & 96.81\% &7.1e-7\\
\hline
2 & 96.38\% & 95.18\% & 95.82\% &9.8e-6\\
\hline
3 & 92.21\% & 78.3\% & 84.78\% &1.7e-3\\
\hline
4 & 62.99\% & 26.33\% & 40.83\% &9.7e-3\\
\hline
6 & 17.88\% & 10.45\% & 13.33\% &5.0e-3\\
\hline
8 & 14.54\% & 8.22\% & 10.84\% &1.8e-4\\
\hline
10 & 11.93\% & 7.91\% & 10.29\% &1.2e-4\\
\hline
\end{tabular}
\end{table}

\subsection{Issues for searching radius}
First of all, we show how the perturbations affect the model performance on Fashion MNIST data in Figure 2. We can see the dimension of 11 has a higher variation. Next is the model with the BN layer performs poorly. This is because we only use the Leakyrelu activation function hence we will not face the vanishing gradients or exploding gradients issue of introducing the BN layer. Our formulation is typically ideal because the gradient is $0$ or $1$. Hence adding the BN layer will only lower the training speed, which is against their original design. Another is BN layer will enlarge the impact of perturbations shown in Figure 2. The same reason is that the beginning variance of processing data is $1$. The variance of learned distribution will decrease if the previous layers work, and transforming them into standard normal distribution will make the scale parameter always larger than $1$. To further validate this, we test the model with different epochs, and the more epochs, the more instability of the results. Last but not the least, the MCR loss seems to have a similar consideration as dropout or BNN, because here we do not use dropout, and we can always get similar results no matter how many epochs we trained.

\begin{figure}[hpbt]
     \centering
         \centering{
         \includegraphics*[width=3cm]{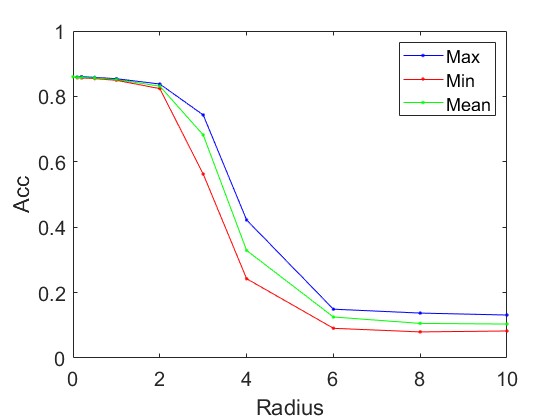}
         \caption*{(a) CNN without bias or BN, 128 dim}
         \includegraphics*[width=3cm]{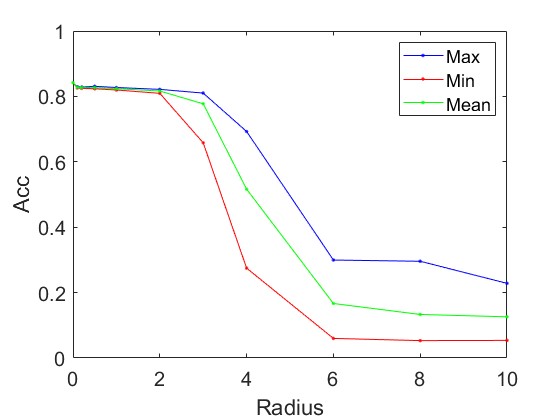}
         \caption*{(b) CNN without bias or BN, 11 dim}
         \includegraphics*[width=3cm]{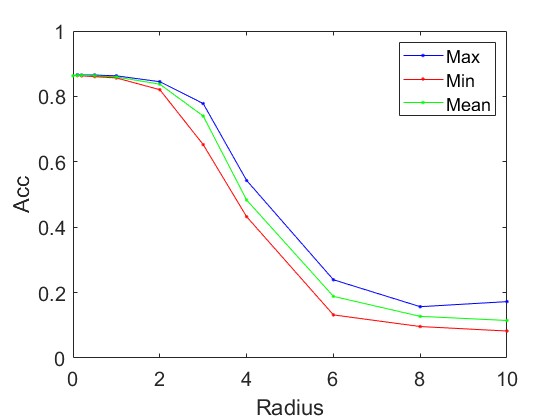}
         \caption*{(c) CNN with bias 128 dim}
         \includegraphics*[width=3cm]{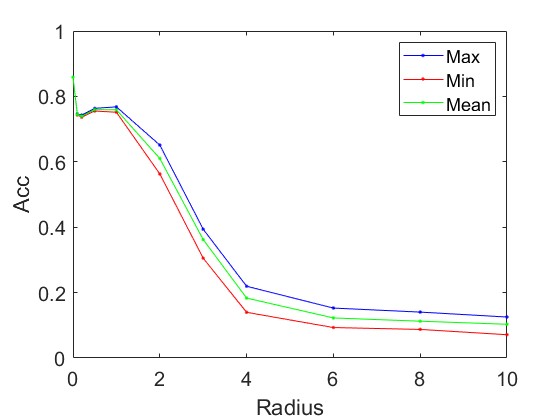}
         \caption*{(d) CNN with BN, 128 dim}}
\caption{Perturbation performance of different models on FMNIST (20 samples, 1000 epochs), noting bias is $0$ to have a fair comparison.}
\label{fig_function2}
\end{figure}


After examining the noise perturbation on the trained neural networks, we check the issues mentioned to find the radius $r$ under the given loss gap requirement. In Figure 3, we can see that enough dimension is critical to search for $r$ this is because the function will be not concave in the larger perturbation area, which we have seen in Figure 1. However, adding bias or the BN layer will also lead to the same situation. People might argue that the selective zone $[0,100]$ is too casual because the preserve area is from $0$ to $10$. The issue is that we do not know exactly what area the zone should be, and we also test for the $log$ case and the normal case with the BN layer from $0$ to $10$. Normally $log$ case can successfully find $r$ while the normal case cannot. Again, setting a suitable zone or loss gap and resampling can effectively find the suitable radius $r$. Lastly, the comparison of the golden search and grid search validates the effectiveness and points out the issues why in other cases we can not find the suitable $r$ with the golden search, see Figure 4,5, and 6.

\begin{figure}[hpbt]
     \centering
         \centering{
         \includegraphics*[width=4cm]{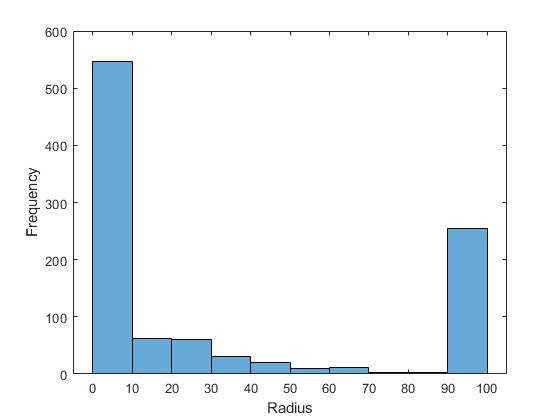}
         \caption*{(a) CNN without bias or BN, 11 dim}
         \includegraphics*[width=4cm]{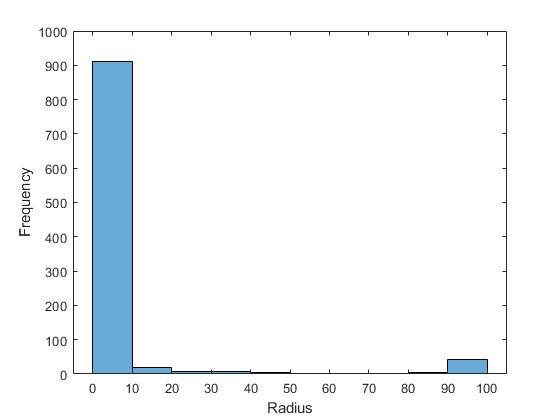}
         \caption*{(b) CNN without bias or BN, 32 dim}
         \includegraphics*[width=4cm]{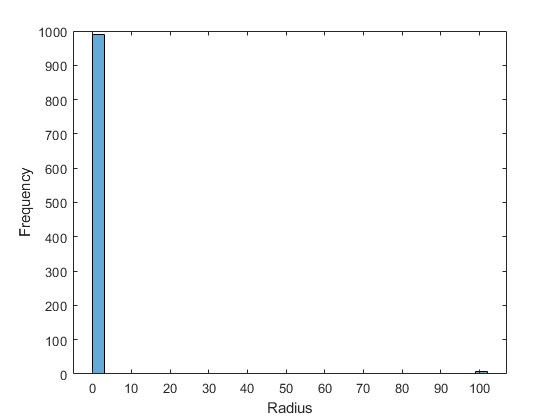}
         \caption*{(c) CNN without bias or BN, 64 dim}
         \includegraphics*[width=4cm]{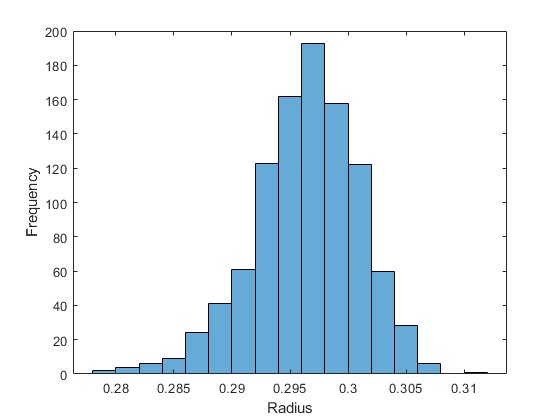}
         \caption*{(d) CNN without bias or BN, 128 dim}
         \includegraphics*[width=4cm]{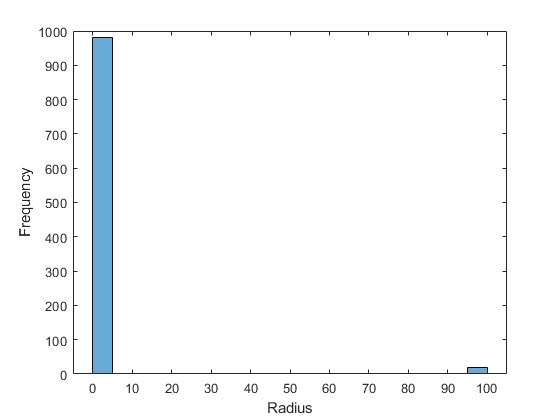}
         \caption*{(e) CNN with bias, 128 dim}
         \includegraphics*[width=4cm]{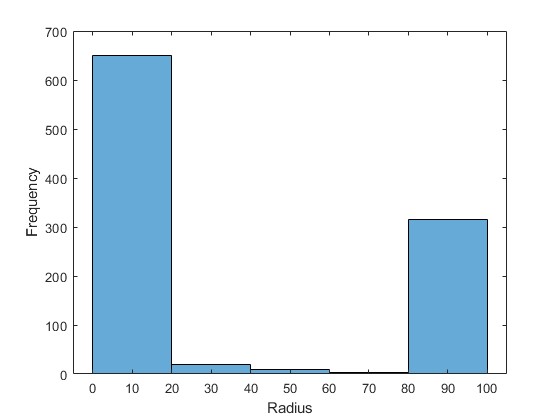}
         \caption*{(f) CNN with BN, 128 dim}}
\caption{Histogram of search $r$ with the golden search on MNIST ($log$ case, 1000 samples from 0 to 100).}
\label{fig_function3}
\end{figure}


\begin{figure}[hpbt]
     \centering
         \centering{
         \includegraphics*[width=3cm]{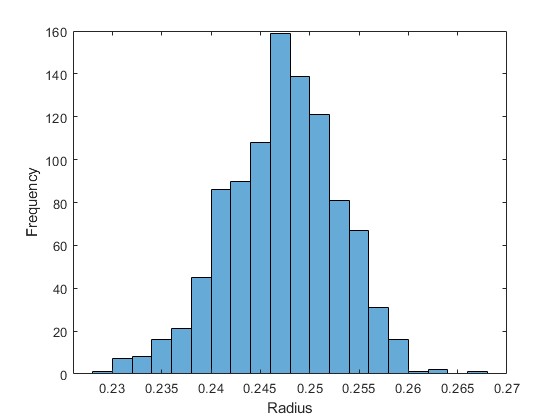}
         \caption*{(a) Golden search, $log$}
         \includegraphics*[width=3cm]{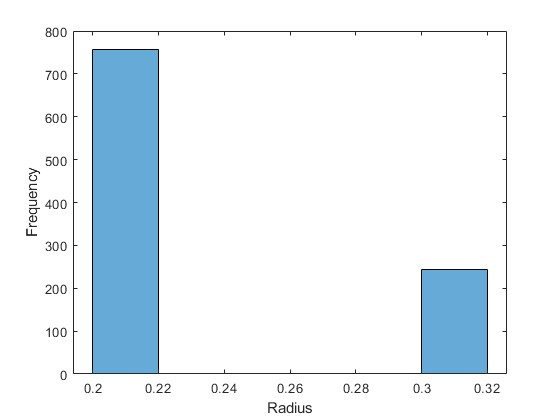}
         \caption*{(b) Grid Search, $log$}
         \includegraphics*[width=3cm]{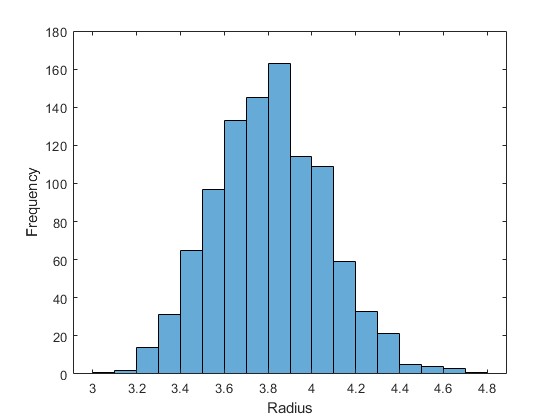}
         \caption*{(c) Golden search}
         \includegraphics*[width=3cm]{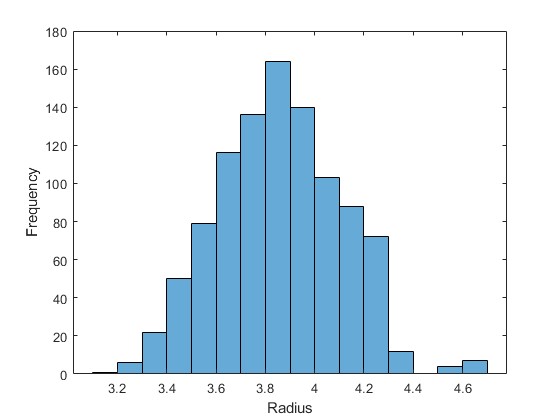}
         \caption*{(d) Golden search}}
\caption{Radius comparison for golden search and grid search for the model without bias on MNIST.}
\label{fig_function4}
\end{figure}


\begin{figure}[hpbt]
     \centering
         \centering{
         \includegraphics*[width=3cm]{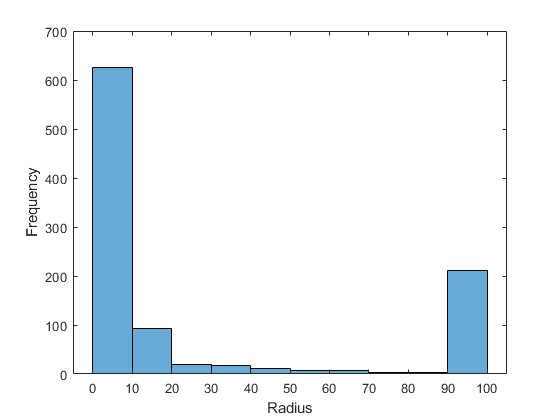}
         \caption*{(a) Golden search, $log$}
         \includegraphics*[width=3cm]{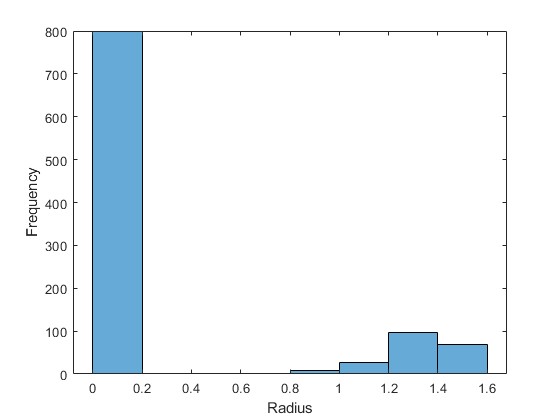}
         \caption*{(b) Grid Search, $log$}}
\caption{Radius comparison for Golden search and Grid search for the models with 11 dimensions on MNIST.}
\label{fig_function5}
\end{figure}


\begin{figure}[hpbt]
     \centering
         \centering{
         \includegraphics*[width=3cm]{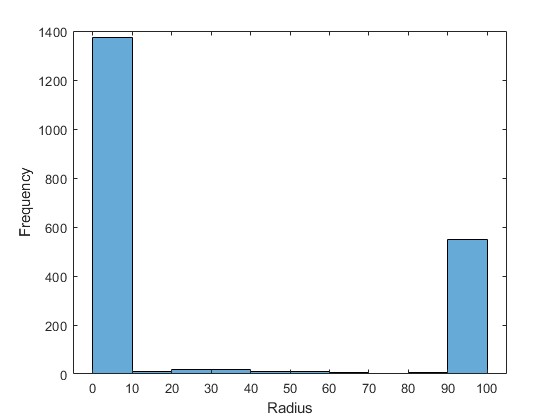}
         \caption*{(a) Golden search, $log$}
         \includegraphics*[width=3cm]{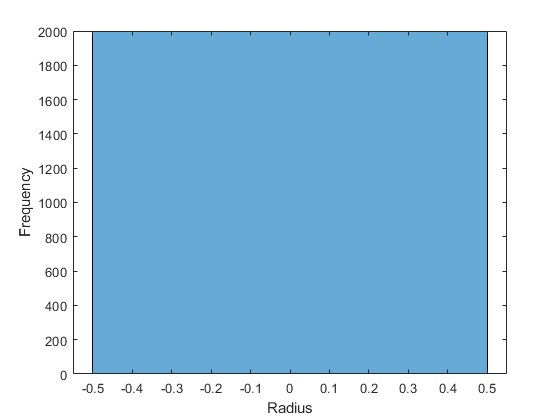}
         \caption*{(b) Grid Search, $log$}
         \includegraphics*[width=3cm]{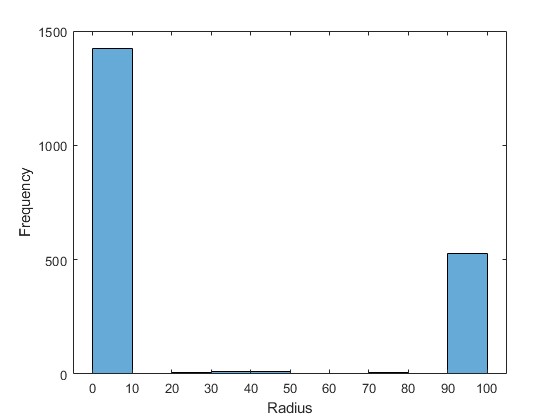}
         \caption*{(c) Golden search}
         \includegraphics*[width=3cm]{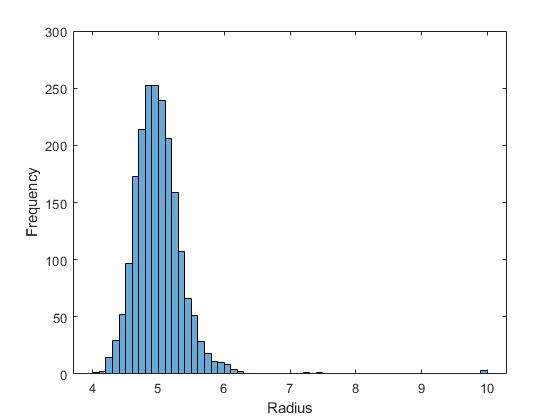}
         \caption*{(d) Golden search}}
\caption{Radius comparison for Golden search and Grid search for the models with BN on MNIST.}
\label{fig_function6}
\end{figure}


\subsection{Noise perturbation}
Furthermore, we test how the $r$ will change when adding noise to the data. In this part, we set the corrupt ratio as $0, 0.05, 0.1, 0.2, 0.3, 0.4, 0.5, 0.6, 0.7, 0.8, 0.9$ with Gaussian random noise with normalization with the data, and search their corresponding radius see Figure 7. For adding small noise, the $r$ will reduce because they are equivalent for small perturbations as $A(X+\xi)+b = (A+\xi)X+b$ with a nonlinear activation function from the Taylor expansion perspective.

\begin{figure}[hpbt]
     \centering
         \centering{
         \includegraphics*[width=3cm]{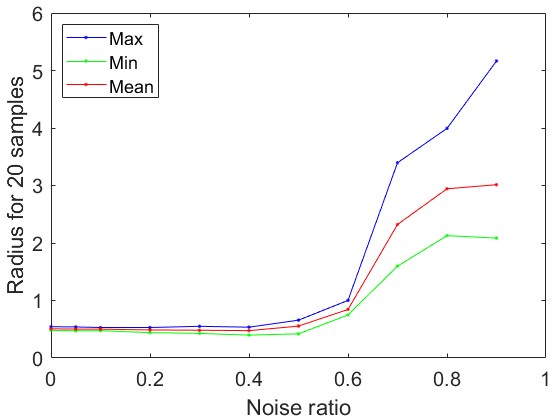}
         \caption*{(a) Corruption in MNIST}
         \includegraphics*[width=3cm]{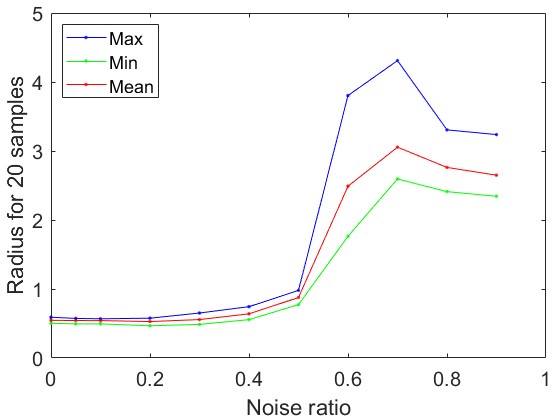}
         \caption*{(b) Corruption in FMNIST}}
\caption{Noise corruption to data (20 samples).}
\label{fig_function7}
\end{figure}


\subsection{Dataset similarity}
Our formulation for out-of-distribution detection is similar to \cite{liang2017enhancing} because $A(X+\xi)+b = (A+\xi)X+b$  as mentioned before. The drawback is that adding noise to the neural network is usually more costly. The flexibility is that we can adjust the noise levels from small to large in contrast to the classical BNN or add noise to data \cite{liang2017enhancing}. This belongs to a simple trick that can be applied to any well-trained neural network. In Tables 2 and 3, we generate the corresponding radius $r$ for different data sets to evaluate their similarity, noting similar results have been given \cite{liang2017enhancing, mukhoti2023deep}.

\begin{table}[!t]
\caption{Different r for other data sets, trained by MNIST}
\centering
\begin{tabular}{|c||c||c||c||c|}
\hline
r ($log$) & Max & Min & mean & var \\
\hline
MNIST & 0.544& 0.481 & 0.507 & 3.1e-4\\
\hline
FMNIST & 1.768 & 1.186 & 1.473 & 0.0184\\
\hline
CIFAR-10(channel 1) & 3.134 & 1.999 & 2.639 &0.102\\
\hline
Gaussian & 3.751 & 1.778 & 2.661 &0.311\\
\hline
Laplace & 5.637 & 2.902 & 3.800 &0.494\\
\hline
Cauchy & 5.807 & 4.188& 4.942 &0.186\\
\hline
\end{tabular}
\end{table}

\begin{table}[!t]
\caption{Different r for other data sets, trained by FMNIST}
\centering
\begin{tabular}{|c||c||c||c||c|}
\hline
r ($log$) & Max & Min & mean & var \\
\hline
FMNIST & 0.591 & 0.503 & 0.546 & 5.4e-4\\
\hline
MNIST & 1.904 & 1.479 & 1.671 & 0.016\\
\hline
CIFAR-10(channel 1) & 2.599 & 2.155 & 2.458 &0.012\\
\hline
Gaussian & 3.220 & 2.189 & 2.560 &0.076\\
\hline
Laplace & 4.129 & 2.852 & 3.487 &0.094\\
\hline
Cauchy & 5.851 & 4.121 & 4.723 &0.161\\
\hline
\end{tabular}
\end{table}

\section{Conclusions and Discussion}
\label{section: conclusion}
In this paper, we apply the minimax game to the classical BNN. From a different perspective, we understand the closed-loop neural networks with the minimax loss and provide a specific explanation for the loss. Close-loop is important while implementing through the minimax coding loss is similar to the classical BNN with a pretty large variance. Compared with classical BNN, this framework is more conservative and is more flexible in adjusting the variance level for any well-trained models. Also, this framework can be used for out-of-distribution scenarios albeit more complicated in contrast to previous methods. Last but not the least, we explain why BN is not suitable in our scenario and summarize the issues for searching $r$ with the golden search method, we believe this is because we can not perfectly transform a distribution to standard normal without error for real experiments.  

\printbibliography

@article{lecun1998mnist,
  title={The MNIST database of handwritten digits},
  author={LeCun, Yann},
  journal={http://yann. lecun. com/exdb/mnist/},
  year={1998}
}

@article{xiao2017fashion,
  title={Fashion-mnist: a novel image dataset for benchmarking machine learning algorithms},
  author={Xiao, Han and Rasul, Kashif and Vollgraf, Roland},
  journal={arXiv preprint arXiv:1708.07747},
  year={2017}
}

@article{krizhevsky2009learning,
  title={Learning multiple layers of features from tiny images},
  author={Krizhevsky, Alex and Hinton, Geoffrey and others},
  year={2009},
  publisher={Toronto, ON, Canada}
}

@inproceedings{gal2016dropout,
  title={Dropout as a bayesian approximation: Representing model uncertainty in deep learning},
  author={Gal, Yarin and Ghahramani, Zoubin},
  booktitle={international conference on machine learning},
  pages={1050--1059},
  year={2016},
  organization={PMLR}
}

@article{duddu2019adversarial,
  title={Adversarial fault tolerant training for deep neural networks},
  author={Duddu, Vasisht and Rao, D Vijay and Balas, Valentina E},
  journal={arXiv preprint arXiv:1907.03103},
  year={2019}
}

@article{neti1992maximally,
  title={Maximally fault tolerant neural networks},
  author={Neti, Chalapathy and Schneider, Michael H and Young, Eric D},
  journal={IEEE Transactions on Neural Networks},
  volume={3},
  number={1},
  pages={14--23},
  year={1992},
  publisher={IEEE}
}

@article{deodhare1998synthesis,
  title={Synthesis of fault-tolerant feedforward neural networks using minimax optimization},
  author={Deodhare, Dipti and Vidyasagar, M and Keethi, S Sathiya},
  journal={IEEE Transactions on Neural Networks},
  volume={9},
  number={5},
  pages={891--900},
  year={1998},
  publisher={IEEE}
}

@inproceedings{liu2018analyzing,
  title={Analyzing the noise robustness of deep neural networks},
  author={Liu, Mengchen and Liu, Shixia and Su, Hang and Cao, Kelei and Zhu, Jun},
  booktitle={2018 IEEE Conference on Visual Analytics Science and Technology (VAST)},
  pages={60--71},
  year={2018},
  organization={IEEE}
}

@article{fawzi2017robustness,
  title={The robustness of deep networks: A geometrical perspective},
  author={Fawzi, Alhussein and Moosavi-Dezfooli, Seyed-Mohsen and Frossard, Pascal},
  journal={IEEE Signal Processing Magazine},
  volume={34},
  number={6},
  pages={50--62},
  year={2017},
  publisher={IEEE}
}

@article{malinin2018predictive,
  title={Predictive uncertainty estimation via prior networks},
  author={Malinin, Andrey and Gales, Mark},
  journal={Advances in neural information processing systems},
  volume={31},
  year={2018}
}

@article{jospin2022hands,
  title={Hands-on Bayesian neural networks—A tutorial for deep learning users},
  author={Jospin, Laurent Valentin and Laga, Hamid and Boussaid, Farid and Buntine, Wray and Bennamoun, Mohammed},
  journal={IEEE Computational Intelligence Magazine},
  volume={17},
  number={2},
  pages={29--48},
  year={2022},
  publisher={IEEE}
}

@article{dai2022ctrl,
  title={Ctrl: Closed-loop transcription to an ldr via minimaxing rate reduction},
  author={Dai, Xili and Tong, Shengbang and Li, Mingyang and Wu, Ziyang and Psenka, Michael and Chan, Kwan Ho Ryan and Zhai, Pengyuan and Yu, Yaodong and Yuan, Xiaojun and Shum, Heung-Yeung and others},
  journal={Entropy},
  volume={24},
  number={4},
  pages={456},
  year={2022},
  publisher={MDPI}
}

@article{dai2023closed,
  title={Closed-Loop Transcription via Convolutional Sparse Coding},
  author={Dai, Xili and Chen, Ke and Tong, Shengbang and Zhang, Jingyuan and Gao, Xingjian and Li, Mingyang and Pai, Druv and Zhai, Yuexiang and Yuan, XIaojun and Shum, Heung-Yeung and others},
  journal={arXiv preprint arXiv:2302.09347},
  year={2023}
}

@article{yu2020learning,
  title={Learning diverse and discriminative representations via the principle of maximal coding rate reduction},
  author={Yu, Yaodong and Chan, Kwan Ho Ryan and You, Chong and Song, Chaobing and Ma, Yi},
  journal={Advances in Neural Information Processing Systems},
  volume={33},
  pages={9422--9434},
  year={2020}
}

@inproceedings{blundell2015weight,
  title={Weight uncertainty in neural network},
  author={Blundell, Charles and Cornebise, Julien and Kavukcuoglu, Koray and Wierstra, Daan},
  booktitle={International conference on machine learning},
  pages={1613--1622},
  year={2015},
  organization={PMLR}
}

@article{kramer2016scikit,
  title={Scikit-learn},
  author={Kramer, Oliver and Kramer, Oliver},
  journal={Machine learning for evolution strategies},
  pages={45--53},
  year={2016},
  publisher={Springer}
}

@inproceedings{guo2003knn,
  title={KNN model-based approach in classification},
  author={Guo, Gongde and Wang, Hui and Bell, David and Bi, Yaxin and Greer, Kieran},
  booktitle={On The Move to Meaningful Internet Systems 2003: CoopIS, DOA, and ODBASE: OTM Confederated International Conferences, CoopIS, DOA, and ODBASE 2003, Catania, Sicily, Italy, November 3-7, 2003. Proceedings},
  pages={986--996},
  year={2003},
  organization={Springer}
}

@article{carbone2020robustness,
  title={Robustness of bayesian neural networks to gradient-based attacks},
  author={Carbone, Ginevra and Wicker, Matthew and Laurenti, Luca and Patane, Andrea and Bortolussi, Luca and Sanguinetti, Guido},
  journal={Advances in Neural Information Processing Systems},
  volume={33},
  pages={15602--15613},
  year={2020}
}

@article{bi2022pangu,
  title={Pangu-weather: A 3d high-resolution model for fast and accurate global weather forecast},
  author={Bi, Kaifeng and Xie, Lingxi and Zhang, Hengheng and Chen, Xin and Gu, Xiaotao and Tian, Qi},
  journal={arXiv preprint arXiv:2211.02556},
  year={2022}
}

@article{wei2022emergent,
  title={Emergent abilities of large language models},
  author={Wei, Jason and Tay, Yi and Bommasani, Rishi and Raffel, Colin and Zoph, Barret and Borgeaud, Sebastian and Yogatama, Dani and Bosma, Maarten and Zhou, Denny and Metzler, Donald and others},
  journal={arXiv preprint arXiv:2206.07682},
  year={2022}
}

@article{li2019deep,
  title={Deep learning for hyperspectral image classification: An overview},
  author={Li, Shutao and Song, Weiwei and Fang, Leyuan and Chen, Yushi and Ghamisi, Pedram and Benediktsson, Jon Atli},
  journal={IEEE Transactions on Geoscience and Remote Sensing},
  volume={57},
  number={9},
  pages={6690--6709},
  year={2019},
  publisher={IEEE}
}

@article{paszke2019pytorch,
  title={Pytorch: An imperative style, high-performance deep learning library},
  author={Paszke, Adam and Gross, Sam and Massa, Francisco and Lerer, Adam and Bradbury, James and Chanan, Gregory and Killeen, Trevor and Lin, Zeming and Gimelshein, Natalia and Antiga, Luca and others},
  journal={Advances in neural information processing systems},
  volume={32},
  year={2019}
}

@book{berger2013statistical,
  title={Statistical decision theory and Bayesian analysis},
  author={Berger, James O},
  year={2013},
  publisher={Springer Science \& Business Media}
}

@inproceedings{izmailov2021bayesian,
  title={What are Bayesian neural network posteriors really like?},
  author={Izmailov, Pavel and Vikram, Sharad and Hoffman, Matthew D and Wilson, Andrew Gordon Gordon},
  booktitle={International conference on machine learning},
  pages={4629--4640},
  year={2021},
  organization={PMLR}
}

@article{srivastava2014dropout,
  title={Dropout: a simple way to prevent neural networks from overfitting},
  author={Srivastava, Nitish and Hinton, Geoffrey and Krizhevsky, Alex and Sutskever, Ilya and Salakhutdinov, Ruslan},
  journal={The journal of machine learning research},
  volume={15},
  number={1},
  pages={1929--1958},
  year={2014},
  publisher={JMLR. org}
}

@inproceedings{mukhoti2023deep,
  title={Deep deterministic uncertainty: A new simple baseline},
  author={Mukhoti, Jishnu and Kirsch, Andreas and van Amersfoort, Joost and Torr, Philip HS and Gal, Yarin},
  booktitle={Proceedings of the IEEE/CVF Conference on Computer Vision and Pattern Recognition},
  pages={24384--24394},
  year={2023}
}

@article{hjelm2018learning,
  title={Learning deep representations by mutual information estimation and maximization},
  author={Hjelm, R Devon and Fedorov, Alex and Lavoie-Marchildon, Samuel and Grewal, Karan and Bachman, Phil and Trischler, Adam and Bengio, Yoshua},
  journal={arXiv preprint arXiv:1808.06670},
  year={2018}
}

@article{ioffe2015batch,
  title={Batch normalization: Accelerating deep network training by reducing internal covariate shift},
  author={Ioffe, Sergey},
  journal={arXiv preprint arXiv:1502.03167},
  year={2015}
}

@article{lecun1995convolutional,
  title={Convolutional networks for images, speech, and time series},
  author={LeCun, Yann and Bengio, Yoshua and others},
  journal={The handbook of brain theory and neural networks},
  volume={3361},
  number={10},
  pages={1995},
  year={1995},
  publisher={Citeseer}
}

@article{liang2017enhancing,
  title={Enhancing the reliability of out-of-distribution image detection in neural networks},
  author={Liang, Shiyu and Li, Yixuan and Srikant, Rayadurgam},
  journal={arXiv preprint arXiv:1706.02690},
  year={2017}
}

@book{neal2012bayesian,
  title={Bayesian learning for neural networks},
  author={Neal, Radford M},
  volume={118},
  year={2012},
  publisher={Springer Science \& Business Media}
}

@article{mackay1992practical,
  title={A practical Bayesian framework for backpropagation networks},
  author={MacKay, David JC},
  journal={Neural computation},
  volume={4},
  number={3},
  pages={448--472},
  year={1992},
  publisher={MIT Press One Rogers Street, Cambridge, MA 02142-1209, USA journals-info~…}
}

@inproceedings{buening2023minimax,
  title={Minimax-bayes reinforcement learning},
  author={Buening, Thomas Kleine and Dimitrakakis, Christos and Eriksson, Hannes and Grover, Divya and Jorge, Emilio},
  booktitle={International Conference on Artificial Intelligence and Statistics},
  pages={7511--7527},
  year={2023},
  organization={PMLR}
}

@article{wang2021pac,
  title={Pac-bayes information bottleneck},
  author={Wang, Zifeng and Huang, Shao-Lun and Kuruoglu, Ercan E and Sun, Jimeng and Chen, Xi and Zheng, Yefeng},
  journal={arXiv preprint arXiv:2109.14509},
  year={2021}
}

@misc{kuruoglu2010using,
  title={Using annotations for summarizing a document image and itemizing the summary based on similar annotations},
  author={Kuruoglu, Ercan E and Taylor, Alex S},
  year={2010},
  month=may # "~4",
  publisher={Google Patents},
  note={US Patent 7,712,028}
}


\end{document}